\documentclass[letterpaper, 10 pt, conference]{ieeeconf}  

\IEEEoverridecommandlockouts                              

\usepackage{verbatim}
\usepackage{tcolorbox}
\usepackage{amsmath}
\usepackage{amssymb}
\usepackage{algorithm}
\usepackage{graphicx}
\usepackage[noend]{algpseudocode}
\usepackage{drum}
\usepackage{times}
\usepackage{sidecap}
\usepackage{url}

\usepackage{adjustbox}
\usepackage{multirow}
\tcbuselibrary{most}

\usepackage{multicol}

\newcommand{\software}[1]{\textsf{{\footnotesize\emph{#1}}}}

\newenvironment{tcfigure}[1]
{\stepcounter{figure}%
\tcbset{enhanced,attach boxed title to top center={yshift=-3mm,yshifttext=-1mm},
  colback=white,colbacktitle=black,fonttitle=\bfseries,
  boxed title style={size=small,colframe=black},title=Figure~\thefigure: #1} \begin{tcolorbox}\addtocounter{figure}{-1}}
{\end{tcolorbox}}

\usepackage{etoolbox}
\usepackage{xcolor}
\makeatletter
\patchcmd{\@setref}{\bfseries ??}{\colorbox{red}{?reference?}}{}{}
\patchcmd{\@citex}{\bfseries ?}{\colorbox{red}{?citation?}}{}{}
\makeatother
\graphicspath{{figures/}}

\title{\LARGE \bf
Particle Traces for Detecting Divergent Robot Behavior
}

\author{Samuel Zapolsky$^{*}$ and Evan Drumwright$^{\dag}$
\thanks{$^{*}$Samuel Zapolsky with the Computer Science Dept., George Washington University, Washington, DC, USA     {\tt\small samzapo@gwu.edu}}%
\thanks{$^{\dag}$Evan Drumwright is faculty in the Computer Science Dept., George Washington University, Washington, DC, USA {\tt\small drum@gwu.edu}}%
}

\begin{document}

\maketitle
\thispagestyle{empty}
\pagestyle{empty}


\begin{abstract}
The motion of robots and objects in our world is often highly dependent upon contact. When contact is expected but does not occur or when contact is not expected but does occur, robot behavior diverges from plan, often disastrously. This paper describes an approach that uses simulation to detect possible such behavioral divergences on real robots. This approach, and others like it, could be applied to validation of robot behaviors, mechanism design, and even online planning.

The \emph{particle trace} approach samples robot modeling parameters, sensory readings, and state estimates to evaluate a robot's behavior statistically over a range of conditions.  We demonstrate that combining even coarse estimates of state and modeling parameters with fast multibody simulation can be sufficient to detect divergent robot behavior and characterize robot performance in the real world. Correspondingly, this approach could be used to assess risk and find and analyze likely failures, given the extensive data that such simulations can generate. 

  We assess this approach on actuated, high degree-of-freedom robot locomotion examples, a picking task with a fixed-base manipulator, and an unpowered passive dynamic walker.  This research works toward understanding how multi-rigid body simulations can better characterize the behavior of robots without significantly compliant elements.
\end{abstract}

\IEEEpeerreviewmaketitle

\section{INTRODUCTION}
\label{section:intro}

The standard approach to validating robot behavior is simulated testing followed by \emph{in situ} testing. This approach does not inspire confidence as simulations often fail to reflect real world behavior and \emph{in situ} testing is tedious and slow. This problem has instigated research into formal verification methods for robotics (e.g.,~\cite{Johnson:2015,Posa:2015}), which appears promising; intense study is currently underway to scale these approaches to higher degree of freedom systems. This paper explores an alternative path that is straightforward, easily implemented, and uses techniques already familiar to many roboticists to bridge the extremes of isolated physical simulation tests and full-on testing on real robotic hardware. 

\begin{figure}
\begin{center}
\includegraphics[width=\linewidth]{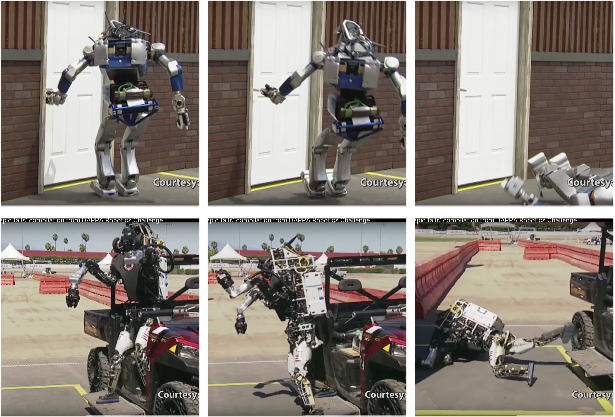}

\end{center}
\caption{\label{fig:expectation} Two robots performing tasks with anticipated contact (images captured from a video of DARPA's recent robotics challenge).  Both control strategies---(top) turning a door handle, (bottom) standing up from a seat---quickly diverge from plan without the anticipated contact, and the robots fall catastrophically. The particle trace approach can be applied to identify brittle aspects of a plan.}  
\end{figure}

Our approach focuses specifically on robots that physically interact with their environment via contact (i.e., manipulation and locomotion). Contact is a governing factor for the movement of legged robots about their environment and for the manner in which robot hands pick up, move, operate, and otherwise manipulate objects in their environment.  As the photos in Figure~\ref{fig:expectation} depict, the unexpected presence or absence of contact can cause catastrophic failure.\footnote{Russ Tedrake claimed that this problem was a dominant cause of failure of the robots in DARPA's Robotics Challenge in a plenary session at Humanoids 2015.}  We seek to iteratively improve control policy and physical design robustness by identifying and modifying policies and designs that are sensitive to modeling and estimation errors.  We perform this task by detecting and addressing novel, nonsmooth, bifurcating events that appear between simulations of perturbed robot models (henceforth denoted \emph{particle traces}). {\it Our results indicate that novel, divergent behavior can be identified efficiently, at least for some tasks performed by some robots, with even a small number of samples}.  

We assess the particle trace approach using \1 a quadrupedal robot and a physically simulated model of this robot performing locomotion tasks; \2 a virtual manipulator robot performing a picking action; and \3 a physically simulated passive dynamic walker~\cite{Coleman:2001}, for which we assess the walking stability empirically over perturbations in modeling parameters. This last task demonstrates the computational efficiency of the sampling approach---particularly in relation to ongoing related work in assessing stability of hybrid dynamical systems that has proven difficult to scale to higher dimensional systems.  

\section{BACKGROUND AND RELATED WORK}
Our present work considers robot dynamics that are well modeled by rigid bodies and rigid or nearly rigid contact. The approach is not necessarily predicated on these assumptions, but the necessary simulations must be sufficiently fast (likely precluding most deformable body simulations).   

\subsection{Nonsmooth mechanics} 
In addition to the challenges of analyzing nonlinear dynamics stability  (of multi-rigid body systems), the problem discussed in this paper requires consideration of \emph{nonsmooth mechanical systems}~\cite{Brogliato:1996}, for which velocities can change discontinuously due to impacts and even non-impacting contact with Coulomb friction~\cite{Stewart:2000a}.  
Multibody dynamics with rigid contact and Coulomb friction---which captures important stick-slip transitions---can be modeled as a differential algebraic equation (DAE):
\begin{align}
\ddot{\vect{q}} & = \vect{f}(\vect{q}, \dot{\vect{q}}, \vect{u}) \\
\vect{0} & = \vect{\phi}(\vect{q})  \label{eqn:DAE2}
\end{align}
where $\vect{\phi}(.)$ is a set of \emph{active} algebraic constraints, out of $m$ total constraints. Some constraints are always active, like bilateral joint constraints. Other constraints are only active if certain conditions are met; e.g., a contact constraint between two polyhedra would only be active when the bodies are in contact at that point and they would otherwise (i.e., without the constraint in place) interpenetrate at that point:
\begin{align}
\dot{\phi}_i(\vect{q}, \dot{\vect{q}}) = 0 \textrm{ if } \phi_i(\vect{q}) = 0 \textrm{ and } \lambda_i > 0, \textrm{ for } i=1,\ldots,m \label{eqn:DAE1}
\end{align} 
where $\lambda_i$ acts as a Lagrange Multiplier (i.e., it is zero if the constraint is active and non-negative otherwise). These kind of problems can be formalized as a differential complementarity problem~\cite{Pang:2008}.

\subsection{Stability analysis and control of nonsmooth systems}
A number of researchers have studied stability analysis and control of nonsmooth mechanical systems~\cite{Brogliato:1996,Tomlin:2000,Tomlin:2003,Prajna:2007,Prajna:2007a,Leine:2008,Leine:2008a,Papachristodoulou:2009,Posa:2015}, for which hybrid dynamical systems have been a common formal model. These systems have been applied toward the study of walking machines and robots, which we have also used as an illustrative application.  

\subsection{Computer Animation}

\cite{Twigg:2007} uses \emph{visual plausibility}, qualitatively undetectable perturbations to collision parameters, to generate a set of possible ``worlds''. Their idea is essentially the inverse of ours: where
we focus on using various perturbations to a simulation to characterize 
robotic behavior and identify possible divergences, they perturb simulations
to try to find plausible, but low probability events. 

\subsection{Robust Control}

	Robust controllers seek control policies that are effective given bounded errors~\cite{Bemporad:2007}, which generally appear as initial state and control signal perturbations; work in robust model predictive control (MPC) also accounts for error in the system model.  Such systems are tested and/or designed by stress-testing that composes perturbing modeling assumptions on initial state, control signal, environment geometry, or contact data and selecting a controller that performs best across all cases.  Robust control has been used for improving the reliability or reducing uncertainty of locomoting systems~\cite{Wang:2009a, Mombaur:2005, Saglam:2014a, Burden:2015},  effecting grasping behaviors~\cite{Kim:2013,Mahler:2015,Weisz:2012,Zheng:2005}, and planning trajectories that reduce system uncertainty~\cite{Johnson:2016}.  Validating such robust controllers through stress-testing, commonly known as \emph{falsification}, seeks to find counter examples to the robustness claims of a controller~\cite{Branicky:2005,Abbas:2013,Esposito:2005}.

\subsection{Monte Carlo method and particle approaches}
There exists a tenuous relationship between our approach and Monte Carlo and particle-based approaches for state estimation of nonsmooth systems~\cite{Duff:2011,Zhang:2013b,Koval:2013,Li:2015,Li:2015a}.  The extent of the similarity is that both use stochasticity in addition to probability distributions over state to generate time series datasets (\emph{traces}) for each perturbation to dynamical system (\emph{particle}) parameters. 
These works expose the interaction between dynamics and rigid contact mechanics, as developed in theory of linear complementarity systems~\cite{Shen:2005}. 

\section{Approach}

The particle trace approach attempts to locate novel contact events, task failures, and other nonsmooth hybrid state transitions that can have a large impact on system state.   


\label{section:approach}
\subsection{Sampling approach}

Each particle's parameterization incorporates the robot, model, environment, and other simulation features that result from a pseudo-random sampling on each of the uncertain elements of a robotic simulation.  Each particle is then ``traced'' over a user-specified time by simulating the sensing, dynamics, and control policy (see Figure~\ref{fig:sample-parameter}). This paper focuses on multi-rigid body dynamics 
with rigid contact because these models capture the first-order effects 
without excessive parameter tuning and because the models' dynamics can be
simulated orders of magnitude more rapidly than the next more representative
set of models (i.e., deformable bodies). Sensory simulation is limited to 
IMU data in the present work.

\subsection{Generating particles}
 A particle's parameterization determines the evolution of the simulated robotic system and serves as the only cause of diverging behavior between the individual particle traces.  \emph{The perturbation to the ``true'' model is not observed by the robot's planning, control, and state estimation systems, which generally assume a known robot model (henceforth denoted the \emph{expected model}) when, e.g., calculating dynamic and kinematic information.}  
 
\begin{figure}[htpb] 
\vspace{0.1in}
\hspace{-0.02in}\includegraphics[width=\linewidth]{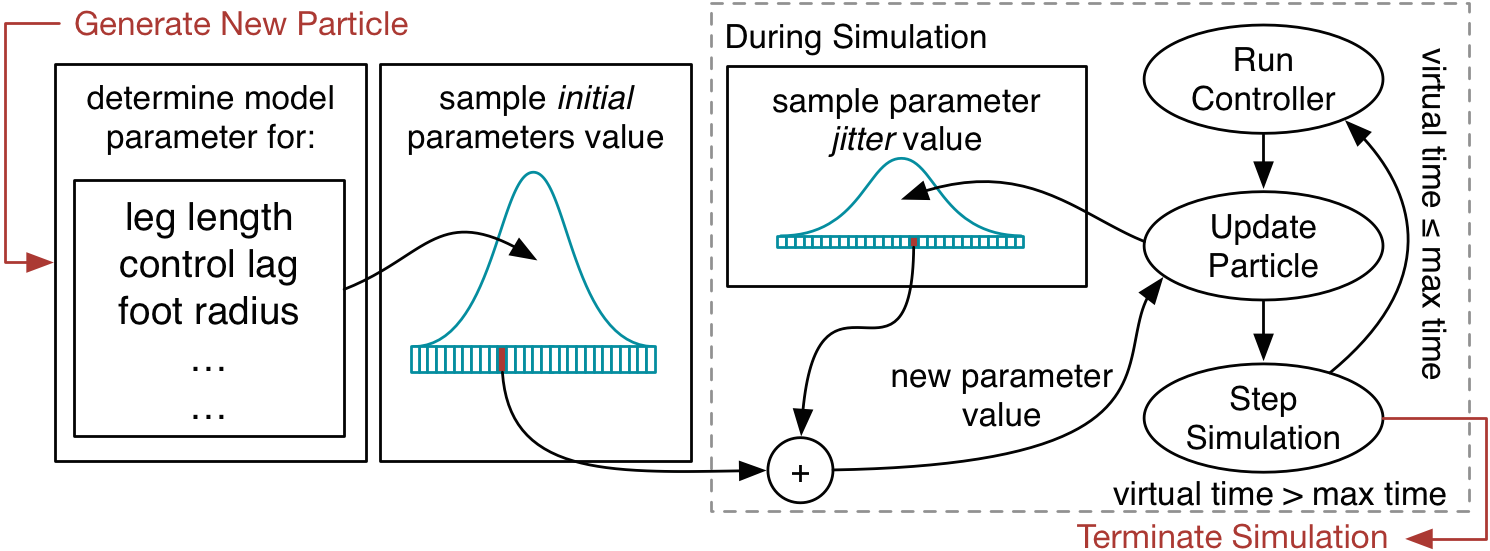}
\caption{\label{fig:sample-parameter} Example of a \emph{particle trace}.  When a new particle is generated:  \1 Initial values of parameters are determined before the first simulation update. \2 Parameters that experience noise online (at each simulation update) are modified from their initial value by sampling a jitter value. } 
\end{figure}

We sample a random perturbation to each particle parameter from an estimated distribution on its uncertainty (usually limiting samples to within three standard deviations, excluding the tails).  Each model and estimation parameter of the particle is then offset from its expected model value using the sampled perturbation \emph{before} starting the simulation.  While a particle is traced as the robot follows its control policy over the course of simulation, additional perturbations to sensor noise and control lag jitter\footnote{Control lag jitter is a small delay that is added/subtracted from the control lag and randomly selected on each control loop iteration.} are sampled \emph{on each control loop iteration}. Figure~\ref{fig:sample-parameter} illustrates this sampling process for a single particle parameter.
 
	We have used normally or uniformly distributed uncertainty on particle parameters (link dimension, link density, joint axes, control lag), and homoscedastic (fixed over time), normally distributed uncertainty for parameter noise (contact friction, control lag jitter, sensor data).  The experiments described in \S\ref{section:experiments} used 
Gaussian and uniform distributions over wide ranges (to effect a safety factor), and we did not attempt to tune the distribution parameters.  The number of distributions and parameters (see Figures~3~\&~\ref{fig:links-geometry}) would likely make such tuning infeasible in any case. Further work will be necessary to assess the ramifications of modeling parameters, state estimates, and sensory noise distributions 
that tend to follow heteroscedastic (varying over time), leptokurtic (fat-tailed), or skewed distributions.

\subsection{Computation Time}  Each particle can be integrated stably in the \software{Moby} simulator at approximately real-time speed: each second of time in simulation (virtual time) takes about a second to compute (wall time).  Pseudorandom sampling decouples allows producing any number of particle traces~$s$ in parallel in linear time $\mathcal{O}(\frac{s T}{c m})$ with respect to the number of particle traces ($s$), where $m$ is the real-time factor of the simulation (\mbox{$m > 1$} being faster than real-time), $T$ is the duration of the experiment in virtual time, and $c$ is the number of processor cores available on the machine.  With enough cores ($c \geq s$), this algorithm can run in constant time.

\begin{figure}[h]
\begin{tcfigure}{Sampleable parameters for \software{Links} robot}
\small
\textbf{\emph{(Parameters determined at the start of a particle trace)}}

\textbf{\small Model:}

\footnotesize
link density: \{$1\times$base, $4\times$hip, $4\times$thigh, $4\times$shin, $4\times$foot\} 

link length: \{$4\times$hip, $4\times$thigh, $4\times$shin\} 

link radius: \{$4\times$hip, $4\times$thigh, $4\times$shin, $4\times$foot\} 

joint axis (conical error): 12$\times$actuated joints 

\textbf{\small Environment:}

\footnotesize
surface friction, surface compliance, contact model, surface geometry

\textbf{\small Initial state:}

\footnotesize
$x,y,z, \psi,\phi, \theta, q_1 \cdots q_{12}, \dot{x},\dot{y},\dot{z}, \dot{\psi},\dot{\phi}, \dot{\theta}, \dot{q}_1 \cdots \dot{q}_{12}$

\textbf{\small Other:}

\footnotesize
control lag

\vspace{0.05in}
\hrule
\vspace{0.05in}

{\small
\emph{\textbf{(Parameters determined during particle trace execution)}}
}

\textbf{\small Encoder noise:}
{\footnotesize $q_1 \cdots q_{12}, \dot{q}_1 \cdots \dot{q}_{12}, \ddot{q}_1 \cdots \ddot{q}_{12}$}

\textbf{\small Force sensor noise:} {\footnotesize
$u_1 \cdots u_{12}$}

\textbf{\small IMU noise:} {\footnotesize $\ddot{x},\ddot{y},\ddot{z},\ddot{\psi},\ddot{\phi}, \ddot{\theta}$}

\textbf{\small GPS noise:} {\footnotesize $x,y$ }

\textbf{\small Magnetometer noise:} {\footnotesize $\psi,\phi, \theta$}

\textbf{\small State Estimation noise:} {\footnotesize $z, \dot{x},\dot{y},\dot{z},\dot{\psi},\dot{\phi}, \dot{\theta} $}

\textbf{\small Sensed actuator torque noise:} {\footnotesize $u_{\mathrm{des},1} \cdots u_{\mathrm{des},12}$}

\textbf{\small Other:} {\footnotesize
control lag jitter }
\end{tcfigure}
\addtocounter{figure}{1}
\end{figure}

\begin{figure}[h]
\centering
\vspace{0.1in}
\includegraphics[width=.8\linewidth]{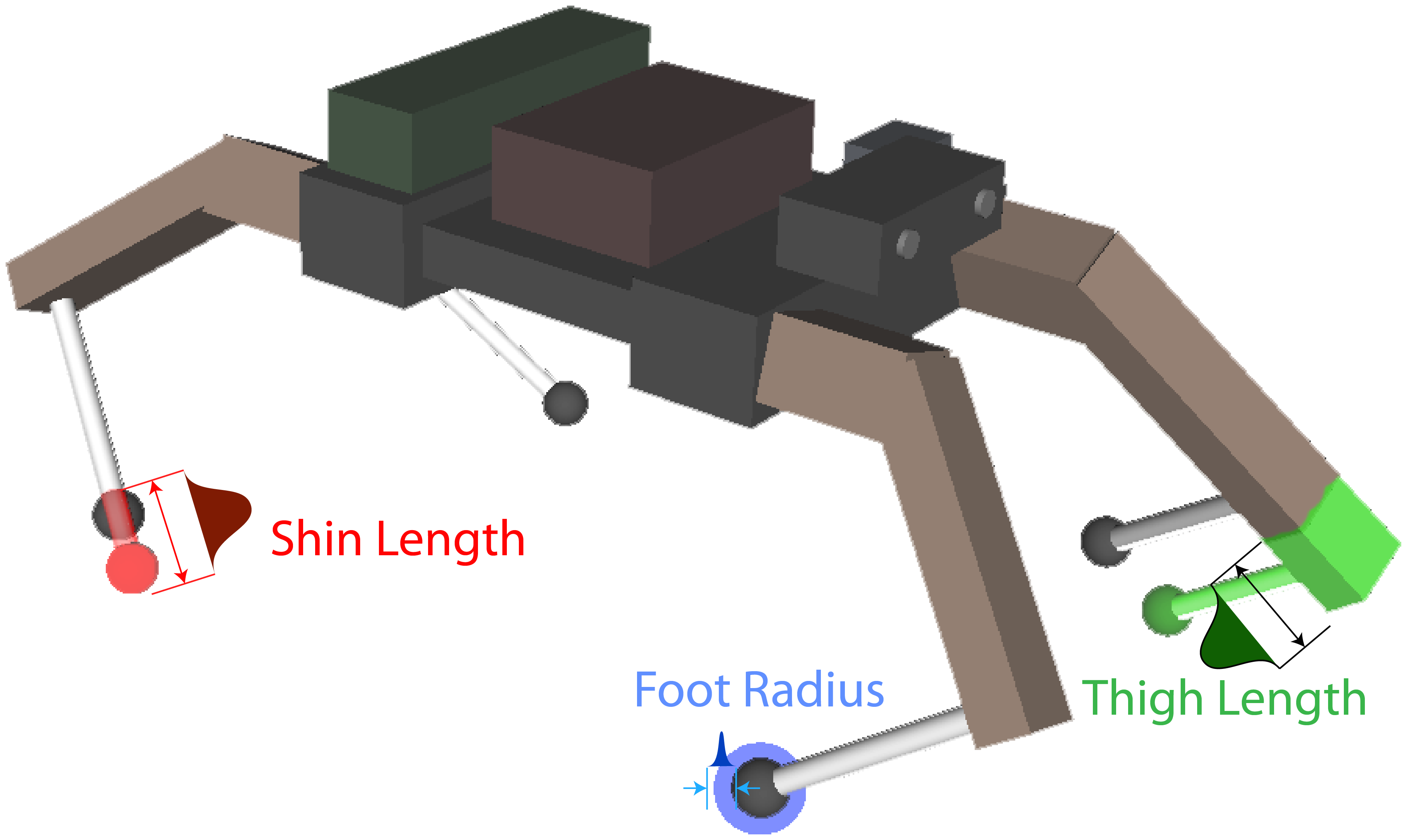}
\caption{\label{fig:links-geometry} A depiction of the probabilistic geometric parameters of a legged robot: shin length, thigh length, and foot radius.}
\end{figure}

\subsection{Identifying divergent behaviors}
\label{section:method:identifying}

We hypothesize that a robot's behavior is likely to be predictable if no nonsmooth events occur or if nonsmooth events occur at similar times between particles.  Similarly, we hypothesize that behavior is harder to predict if some particle traces experience novel nonsmooth events or nonsmooth events occurring in novel sequences.

Accordingly, our approach searches for both ``grazing'' events (events likely to occur in only very particular conditions) and near-miss events. Such events are known in
the hybrid dynamical systems literature as ``grazing bifurcations''~\cite{Budd:1996}. Our second hypothesis anticipates that outcomes will be challenging to predict when a robot operates around such regions of state space. As examples, a slightly longer leg than is modeled might cause the foot to scuff the floor unexpectedly, and a foot heavier than its model might be unable to clear the top of a step when climbing stairs. 

	Given the immense computational resources that can be applied to produce the particle traces, we require a means of identifying divergent  behavior in potentially massive amounts of generated data. We identify divergent behavior automatically by searching for \1 novel events; \2 a novel \emph{sequence} of events; and \3 novel outcome to similar events (between particles). \emph{Event} is used to denote a mode switch, which can occur upon impacts and upon switching between sliding, sticking, and rolling contact. 

Novelty would normally be determined against a baseline, expected behavior.  But often such expectations are hard to predict given a control policy or task description in a complex environment.  Instead, novelty is identified as an event's time, location, or object pair differs from that experienced by other particle traces at a similar time. 

Divergent behavior can also be detected at a goal-oriented level (i.e., rather than detecting novel events) by, e.g., searching for failure to perform a specific task: falls during locomotion, or dropped objects in a pick-and-place task, etc.  Normal behavior can also be determined through consensus, where an outlier would be indicative of divergent behavior.  Our demonstrations in \S\ref{section:experiments} detect divergent behavior through task failure detection.  Further study will focus on efficient detection of detect divergent behavior between particle traces.

\section{Experiments}
\label{section:experiments}

	We assessed the particle trace approach using high-dimensional, non-smooth robotics scenarios. We assess the approach on locomotion scenarios---which feature frequent contact events and probable unexpected, or poorly timed, destabilizing collisions---and a manipulation scenario, because simulation-based plans for grasping have repeatedly proven to be brittle {in situ} (as depicted in Figure~\ref{fig:expectation}). Our experiments use the multi-body dynamics simulator \texttt{Moby}, which has been shown to produce behavior consistent with real robots~\cite{Aukes:2014}, because it uses continuous collision detection~\cite{Mirtich:1996vt} allowing it to locate contact events precisely (see~\cite{Zapolsky:2015}). 

	The following subsections demonstrate how particle traces can be used to determine the sensitivity of a robot to modeling uncertainty (\S\ref{sec:PDwalker}); efficiently locate bifurcating events (\S\ref{sec:bifurcation}); and
assess plan robustness (\S\ref{sec:bifurcation-arm}). \S\ref{section:exp:monte-carlo} shows how the aggregate particle behavior can accurately characterize \emph{in situ} robot behavior.

\subsection{Passive dynamic walker}
\label{sec:PDwalker}

\begin{figure}[htpb]
\centering
\includegraphics[width=0.9\linewidth]{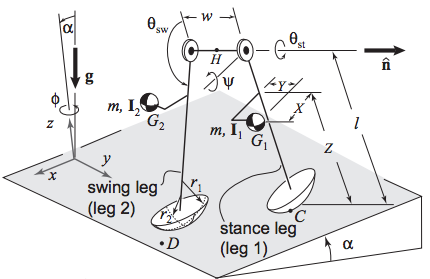}
\caption{\label{fig:walker} Model and state parameters for a passive dynamic walker from Coleman and Ruina~\cite{Coleman:1998}.  This system is stable, but the system's region of stability is small.}
\end{figure}

	We applied the particle trace method to assess the parametric (modeling) limits with respect to stability of a bipedal passive dynamic walker~\cite{Coleman:1998} (illustrated in Figure~\ref{fig:walker}).
 We sampled over the inertial and kinematic parameters of the walker, resulting in four hundred different particles. A fixed point cycle computation process (as described in~\cite{Coleman:1998})
was applied to each particle to compute initial conditions that
yield a walking cycle. Each particle trace was simulated for sufficient duration to allow the biped to walk up to 20 steps.

\begin{figure}[htpb]
\includegraphics[width=\linewidth]{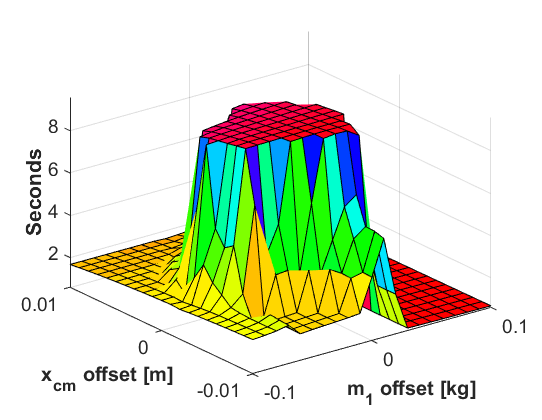}
\caption{\label{fig:walker-data} Stability region for the passive dynamic walker.  Upright time over an 8s walk for varied c.o.m. positions ($x_{\textrm{cm}}$) and foot mass ($m_1$).  $x_{\textrm{cm}}$ is offset from an initial value of 0.0 and $m_1$ is offset from an initial value of 1.0 kg (initially matching $m_2$).}
\end{figure}

All modeling parameters described in Figure~\ref{fig:walker} were perturbed in our assessment: $\{I_{xx}, I_{yy}, I_{zz},I_{xy}, I_{xz}, I_{yz},$ $m_1, \alpha, x_{cm}, y_{cm}, z_{cm} , l, w, r, m_2\} \sim \hspace{-1mm} \mathcal{N}(\mu_\textrm{param},\sigma_\textrm{param})$ where $\mu_\textrm{param}$ is the mean (expected) parameter value for a working simulated system and standard deviation $\sigma_\textrm{param}$ is 5\% of $\mu_\textrm{param}$.  Region(s) of feasible walker parameters can be determined by examining the duration that parameterizations remain upright.

\begin{figure}[htpb]
\includegraphics[width=\linewidth]{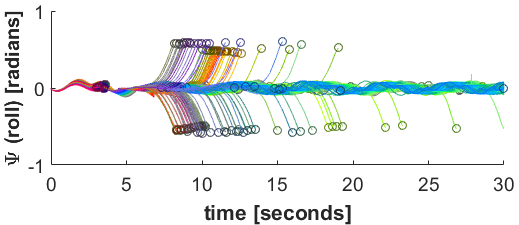}
\caption{\label{fig:walker-data2} 400 randomly sampled particles of the passive walker simulated over 30s (virtual time); the walker's roll orientation is plotted over the course of the experiment.  Termination points (falls) are marked with circles.}
\end{figure}

\textbf{Results:} \quad Figure~\ref{fig:walker-data} depicts the stability region of the passive walker with respect to a grid sampling over the c.o.m. offset (front-back) and left leg mass.  Grid sampling becomes costly when attempted over all parameters due to the curse-of-dimensionality.  A pseudorandom sampling over all fifteen parameters detects similar points of failure for the walking robot, corresponding to the steep ledges seen in Figure~\ref{fig:walker-data}.  Figure~\ref{fig:walker-data2} displays one state value (roll) with respect to time.  State data was collected from four hundred particle traces for this plot.  Falls occur in clusters (at 2s and 8s) as the walker passes through a region of its state space with a grazing bifurcation (usually a scuffed or stubbed foot for this system).

Figure~\ref{fig:walker-data2} depicts a single bifurcating event, scuffing a foot on the ground, generating a compact cluster of state data between particles.  Figure~\ref{fig:walker-data} indicates that a more robust walker can be obtained by increasing $m_1$ slightly ($\approx 0.02$~kg).  

\subsection{Detecting grazing bifurcation for a quadruped}
\label{sec:bifurcation}

\begin{figure}[htpb]
\centering
\vspace{0.1in}
\begin{tabular}{ccc}
\begin{adjustbox}{valign=t}
  \includegraphics[trim={15cm 0cm 15cm 2cm},clip,height=1.57in]{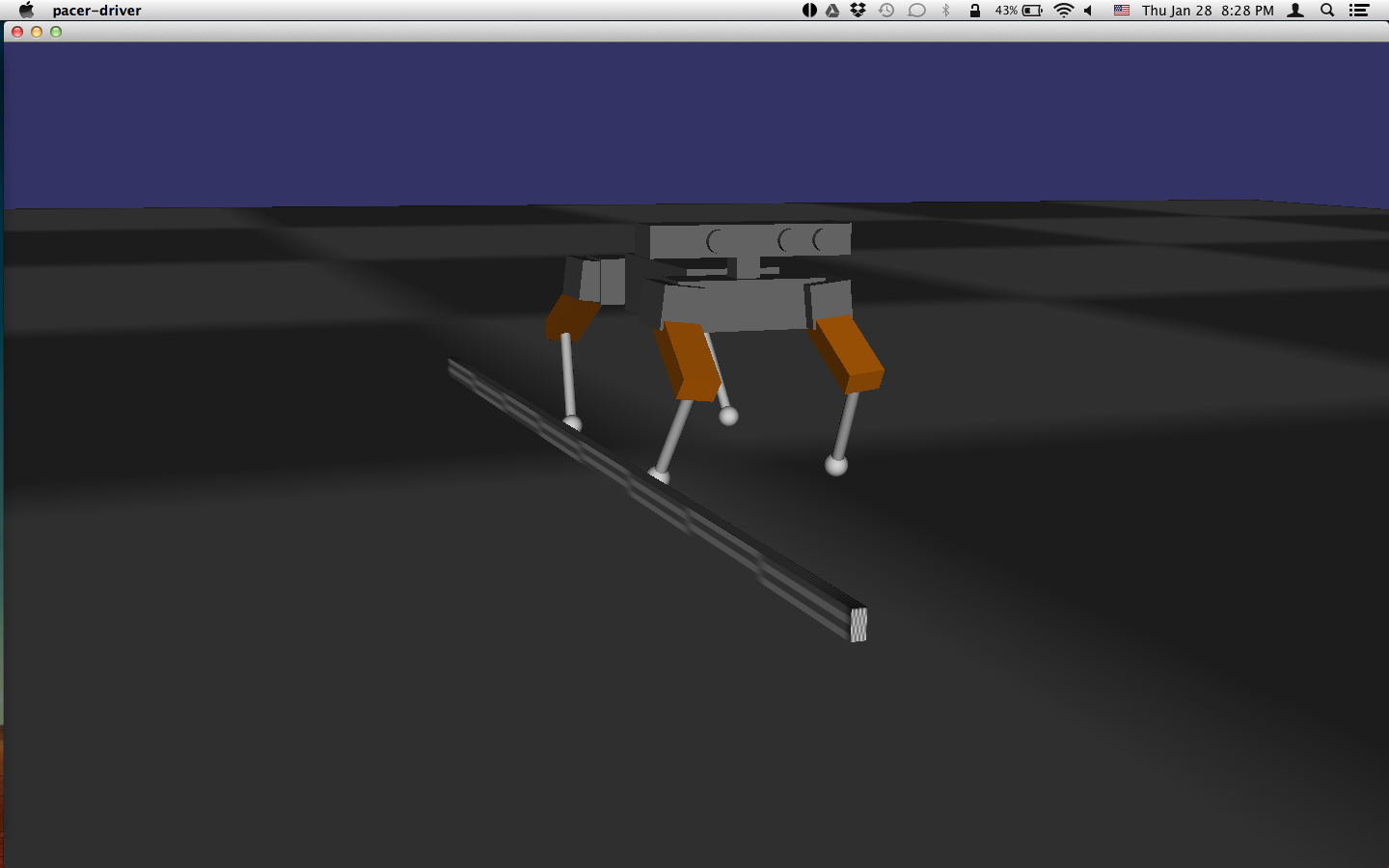}
\end{adjustbox}
&\hspace{-0.15in}
\begin{adjustbox}{valign=t}
\begin{tabular}{@{}c@{}}
\includegraphics[trim={10cm 7cm 12cm 5cm},clip,height=0.75in]{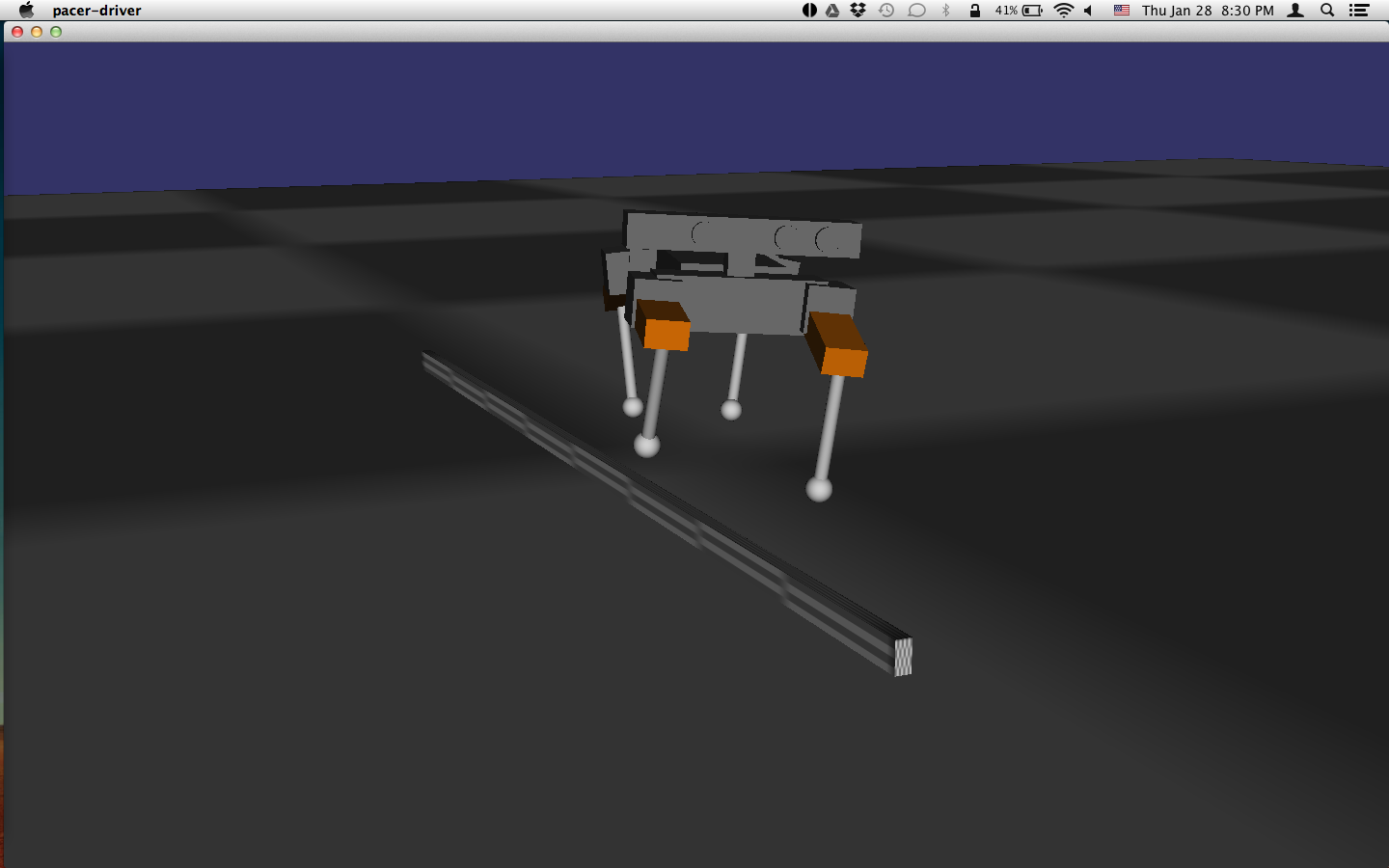} \\[0.2ex] 
  \includegraphics[trim={10cm 7cm 12cm 5cm},clip,height=0.75in]{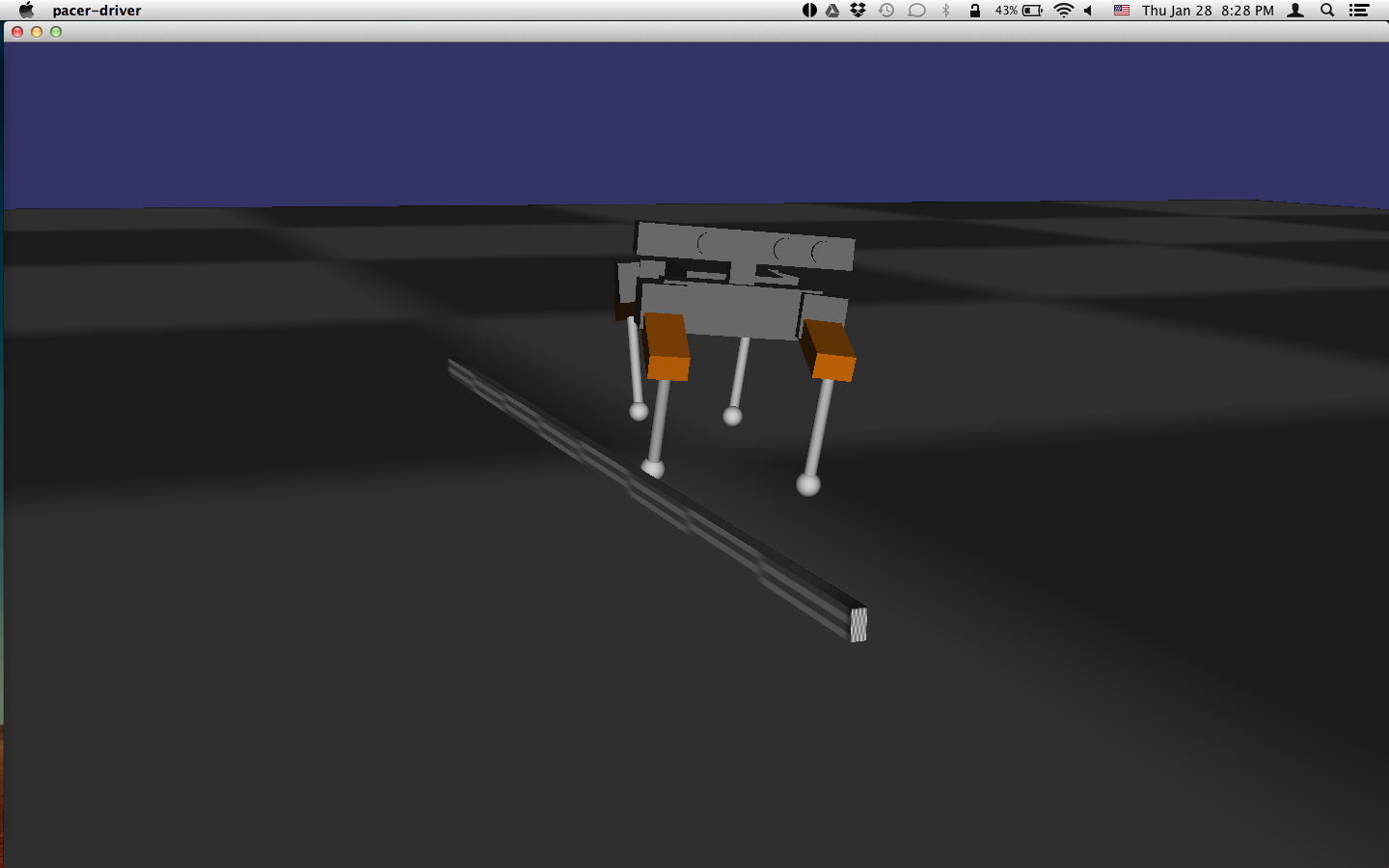}\\
\end{tabular}
\end{adjustbox}
&\hspace{-0.15in}
\begin{adjustbox}{valign=t}
\begin{tabular}{@{}c@{}}
\includegraphics[trim={10cm 7cm 12cm 5cm},clip,height=0.75in]{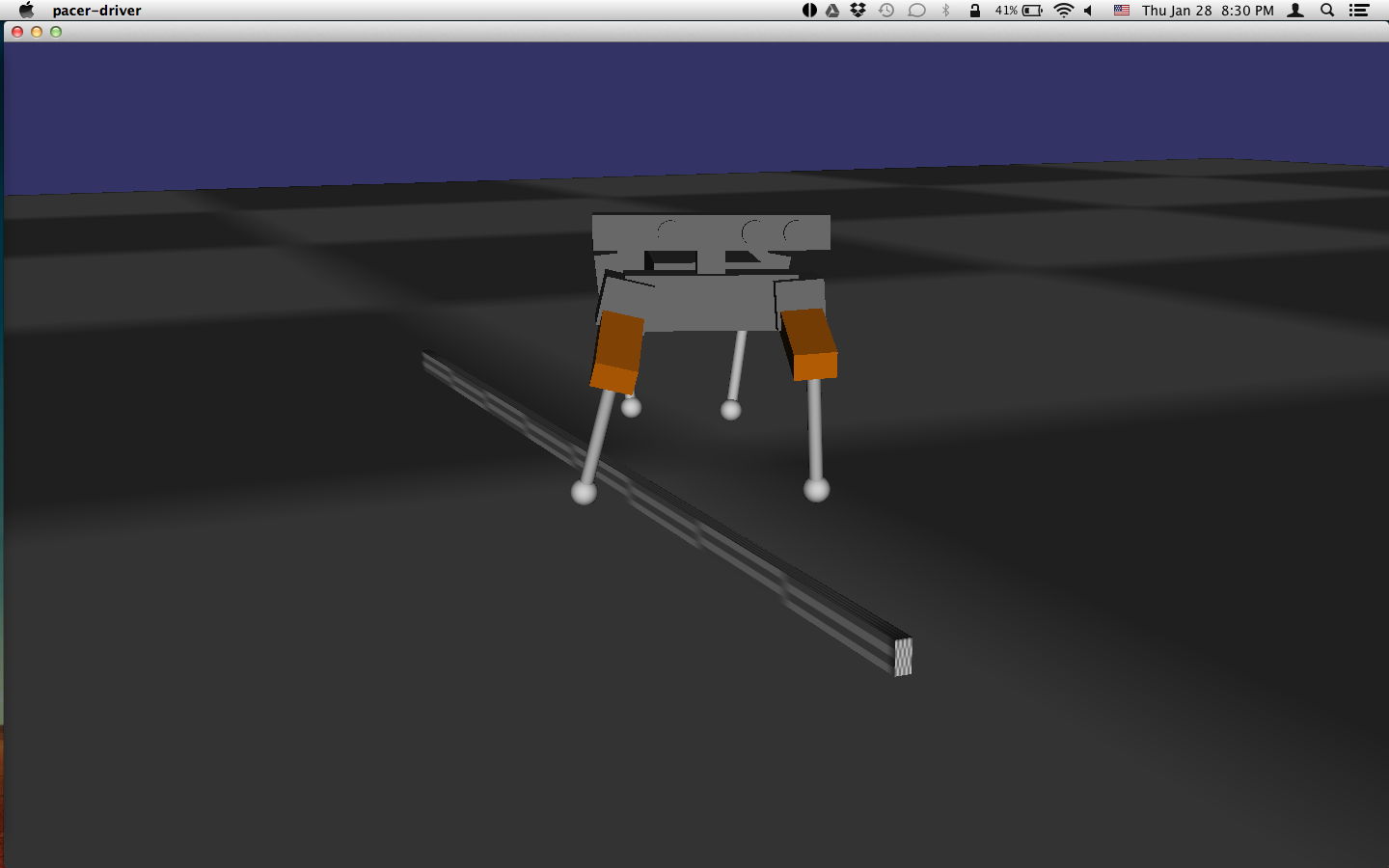}\\[0.2ex]
\includegraphics[trim={10cm 7cm 12cm 5cm},clip,height=0.75in]{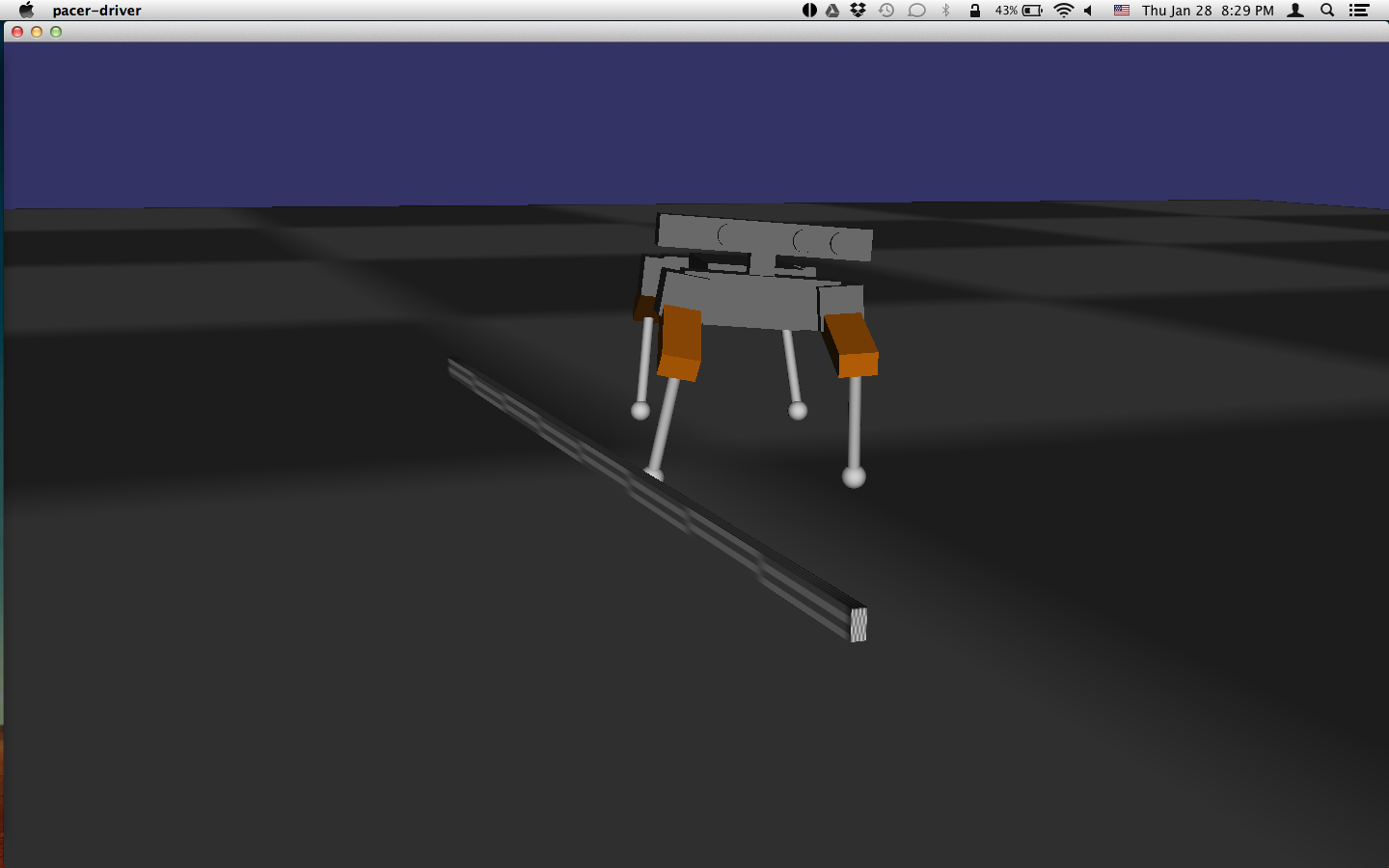}
\end{tabular}
\end{adjustbox}
\end{tabular}
\caption{\label{fig:bifurcation-image}A time-lapse of the virtual \software{Links} robot stepping over (top) or into (bottom) an obstacle given high and low step heights, respectively.}
\end{figure}

We simulated placing a 18 DoF quadrupedal robot model (\emph{Links}) next to a curb obstacle, and directed the quadruped to step over the curb while performing with variable step height (but with parameters otherwise fixed).  A time-lapse depiction of the diverging behavior is shown in Figure~\ref{fig:bifurcation-image}.  The robot clearly will collide with the curb if the robot does not step high enough. We predict that a grazing bifurcation will occur if the step height is approximately equal to the curb height: small changes in initial conditions, modeling parameters, or sensing (of, e.g., curb geometry) will determine whether or not the robot strikes the obstacle.   We ran three trials, each of which uses one of three preset gait control policies that attempt two, three, and four cm step heights.  The curb height was fixed at three cm.

\begin{figure}[H]
\centering
\includegraphics[width=\linewidth]{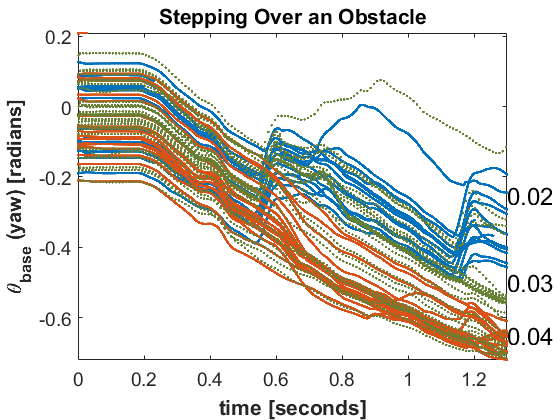}
\caption{\label{fig:links-bifurcation} Virtual quadrupedal robot base yaw when turning into a 3 cm tall curb obstacle.  Each line represents a particle, and each color represents a policy.  \textbf{Red} particles following a 4 cm step height policy (marked as 0.04 m on plot) step over the curb and continue to turn. \textbf{Blue} particles following a 2 cm step height policy (marked as 0.02 m on plot) strike the curb and are prevented from turning.  \textbf{Green, dotted} particles following a 3 cm step height policy (marked as 0.03 m on plot), where step height matches the curb height (3 cm), experience bifurcation by only occasionally striking the obstacle.  The behavior of these particles is less predictable.}
\end{figure}

	\begin{figure*}
\begin{center}
\vspace{0.1in}
\includegraphics[width=0.99\linewidth]{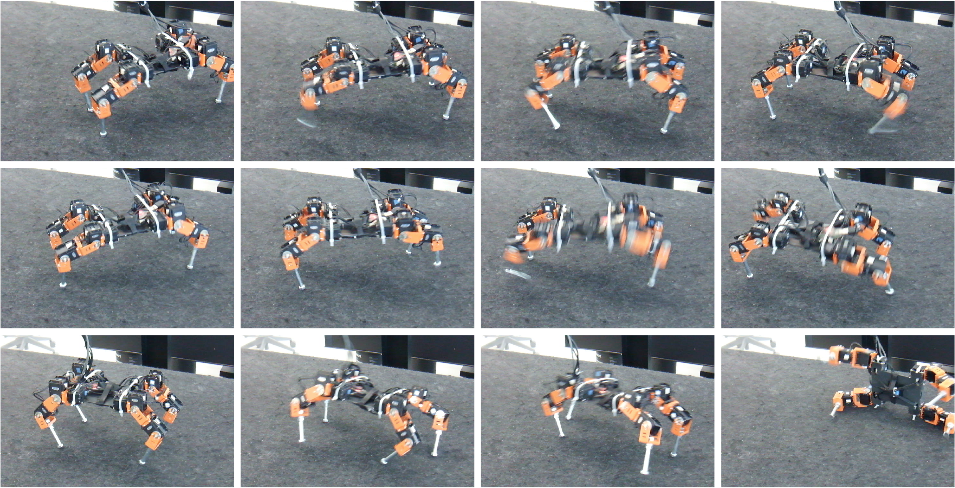}

\end{center}
\vspace{-0.1in}
\caption{\label{fig:links-walking}A two second time-lapse of \software{Links} walking with a gait period duration of: 0.6 seconds (Top); 1.0 seconds (Middle); 1.5 seconds (Bottom).  The robot became progressively less stable as the gait period duration increased.}
\end{figure*}
	 
\textbf{Results:} \ 
Grazing bifurcation drives the traces using the three cm step-height policy into one of two distinct clusters of states that correspond to robot models
clearing and striking the curb, correspondingly  (see Figure~\ref{fig:links-bifurcation}). Comparing the sequence of events in each particle trace for this policy, the cluster that corresponds to robots stepping over the obstacle contains the contact sequence \{~RH/ground, LH/ground, RF/ground, LF/ground~\}, and the cluster that corresponds to robots striking the obstacle contains the contact sequence \{~RH/ground, LH/ground, RF/obstacle, RF/ground, LF/ground~\}. The sampling strategy efficiently uncovers the grazing bifurcation, as the variance of the green paths in Figure~\ref{fig:links-bifurcation} indicate; a grid search over the quadruped's 36 dimensional state space is intractable.  We expect that the robot's \emph{in situ} performance would be difficult to predict under this policy because the robot's behavior is sensitive to modeling and estimation uncertainty. 

In contrast, the four cm step height policy allowed the quadruped to step over the curb for all traces and the two cm step height policy caused the robot to collide with the curb in all traces.  It is reasonable to expect that the real robot would behave predictably under both of these policies. 

\subsection{Assessing grasping plan robustness}
\label{sec:bifurcation-arm}

We simulated an 11 DoF fixed-base manipulator robot performing a picking task (reaching, grasping, and lifting) on a ball within its reach.  Two distinct plans are generated to achieve the tasks: Path A corresponds to the gripper approaching from above the ball, moving in a straight line between the gripper's initial position and the expected position of the ball;  Path B moves sideways from above the ball during the approach to grasp the ball from its side.   We ran two trials, with each trial generating forty particles.  The trials were differentiated only by their approach trajectory and resulting grasp orientation on the ball. 
A trial was considered to be successful if the final position of the ball
matched that planned. We expected that the relative success rates of the
particle traces executing the two plans would reveal the more robust one;
we posit that the absolute success rate of a plan would indicate the
robustness of a plan executed \emph{in situ}, but we did not test this
hypothesis.


\begin{figure}[htpb]
\centering
\includegraphics[width=\linewidth]{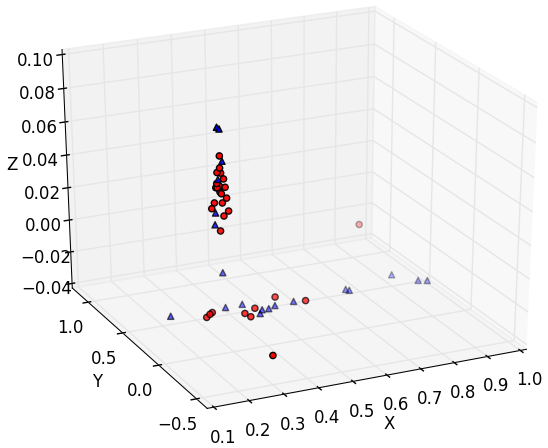}
\caption{\label{fig:final-pos}  Final position of the ball after the pick behavior following Path A (Red Circle) and Path B (Blue Triangle).  Points arrayed along the bottom of the plot are resting on the ground plane.}
\end{figure}

\begin{figure}[htpb]
\centering
\includegraphics[height=2in]{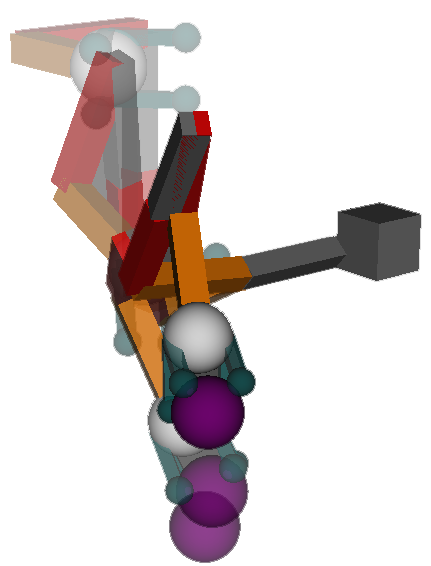}
\includegraphics[height=2in]{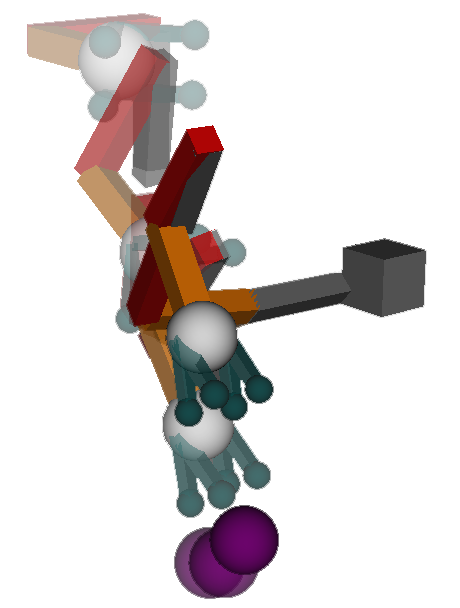}
\caption{\label{fig:plans}  Path A (top) and Path B (bottom) attempting to grasp the ball.  Successes (left) maintain hold on the ball and failures (right) drop the ball or push it away.}
\vspace{0.1in}
\includegraphics[height=2in]{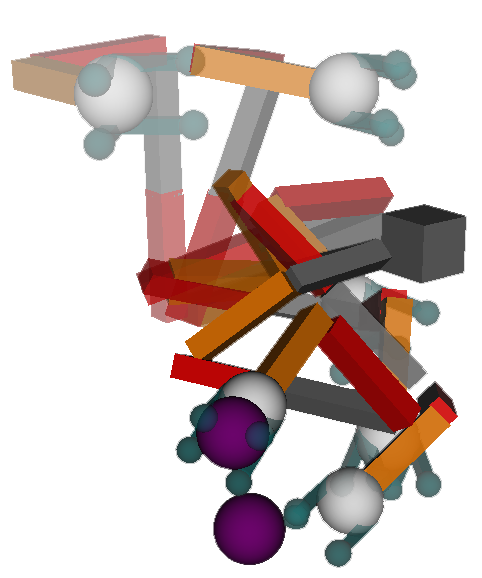}
\includegraphics[height=2in]{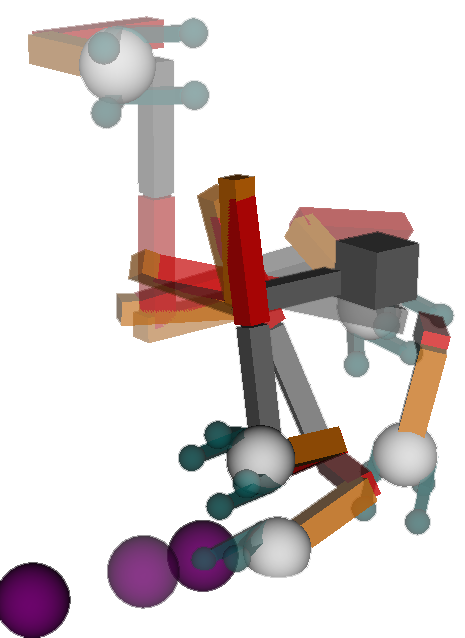}
\end{figure}

\textbf{Results:} \quad Both paths result in the ``original'' (unperturbed model) robot successfully picking the ball from the given initial conditions. These paths could correspond to a brittle plan generated using existing techniques. 
Path A resulted in a 88\% success rate while Path B successfully completed the picking task 63\% (a 40\% performance differential) of the time. A finger tapping the sphere and causing it to roll was a typical cause of failure for Path~B. Figure~\ref{fig:final-pos} shows the final position of the ball in each of the particle traces.  Examples of successful and failing attempts using each trajectory are depicted in Figure~\ref{fig:plans}.  Figure~\ref{fig:plans} depicts a visual realization of what a particle trace might look like within this framework.

\subsection{Statistical behavior of a rigid robot from particle traces}
\label{section:exp:monte-carlo}

The degree of correlation between the behavior of robots simulated over a timespan of seconds or minutes and those robots' physically situated counterparts depends on 
many factors. A flexible robot may evince little of the behavior of its
virtual counterpart simulated using multibody dynamics, for example. This
section focuses on the correlation between simulated and \emph{in situ} behavior for a
scenario that \emph{should} be well modeled using fast simulation tools. This
issue is important because one can only expect grazing bifurcations located
in simulation to be informative if there is some correlation between simulated
and \emph{in situ} behavior. 

\begin{figure}[htpb]
\centering
\includegraphics[height=1.3in]{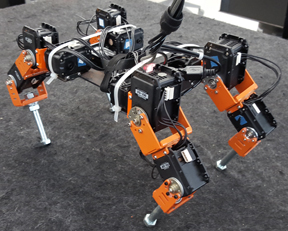}\includegraphics[height=1.3in]{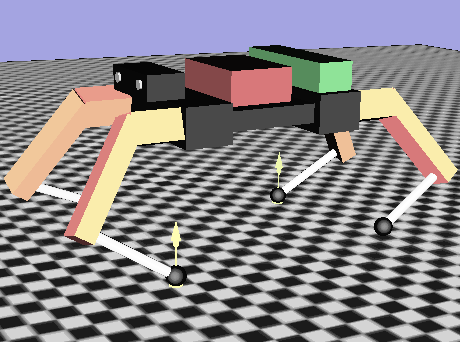}
\caption{\label{fig:links} (Left) the \software{Links} robot in position to begin a walking experiment. (Right) A snapshot of our physical model of \software{Links} in simulation.}
\end{figure}

  \paragraph{Robot} The robot used in physical trials, \software{Links}, is an 18 degree-of-freedom (12 actuated) quadruped robot constructed from Dynamixel actuators and steel links (see Figure~\ref{fig:links}).  Base orientation is recorded by IMU that produces updates at 100 Hz. Modeling uncertainties and errors on even such a small robot are legion and include, but are not limited to, the rigid body assumption, gear backlash, communication delay, IMU sensing delay, the rigid contact assumption, and back EMF. The modeling and state parameters sampled are listed in Figure~3.    

\paragraph{Dynamics model} The virtual quadrupedal robot was modeled using a box geometry for the base link inertia and geometry, cylinders for the limb link inertias and geometries, and spheres for foot inertia and geometries.  Modeling parameters for the virtual quadruped were set from measurements collected from the \software{Links} robot.  There are no compliant elements in the structure of \software{Links} (unless one counts the transmission), allowing it to be readily modeled as a multi-rigid body.

\paragraph{Control Policy} We used a simple gait planning system to move the robot in a walking gait around a one foot diameter circle.  The desired planar motion of the base of the robot $\{\dot{x},\dot{y},\dot{\theta}\}$ is input to the planner, which produces a trajectory for each foot that attempts to drive the robot base toward the intended operational-space configuration while maintaining sticking contact with the ground.  

We adjusted a single gait parameter (gait period duration) and observed how it affected the behavior of the robot.  Gait period duration was adjusted from an empirically observed stable value of 0.6 seconds per gait cycle, upward to a value where we had previously observed definite failure: 1.5 seconds per gait cycle.  Each particle was traced over 20s of virtual time or until a fall, and \software{Links} was permitted to walk for 20s of wall time.  

\begin{figure}[htpb]
\centering
\includegraphics[width=\linewidth]{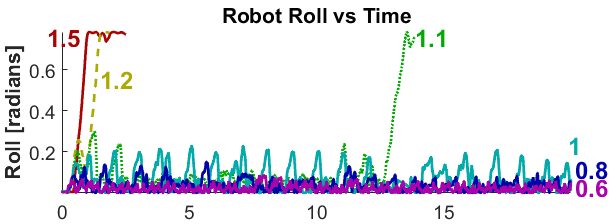}
\includegraphics[width=\linewidth]{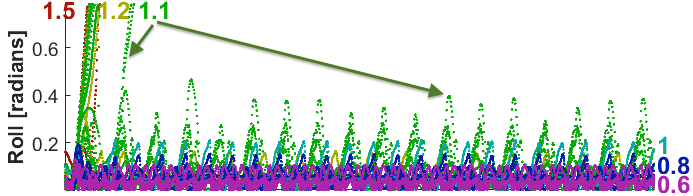}
\caption{\label{fig:links-data} Roll orientation data for the \software{Links} robot walking in a circle: (Top) \emph{in situ} and (Bottom) \emph{in sim}.  Each line is labeled with its corresponding value for the gait period duration).}
\end{figure}

\begin{figure}[htpb]
\includegraphics[width=\linewidth]{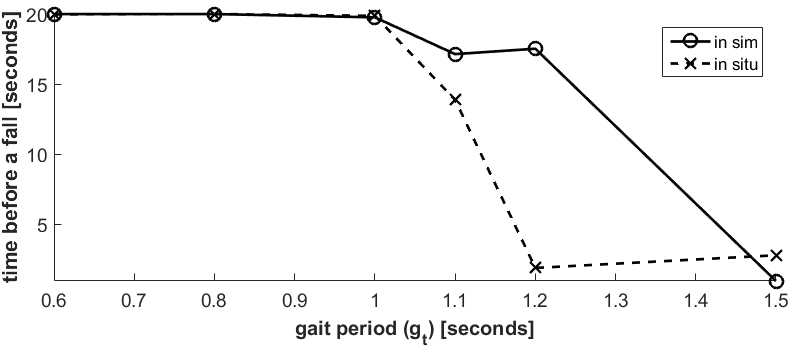}
\caption{\label{fig:links-time}  Duration of time until a fall of the locomoting robot plotted with respect to the gait period duration parameter. ($\times$ mark, dotted line) Duration of wall time until a fall of the \software{Links} robot.  ($\circ$ mark, solid line) Average duration of virtual time until a fall for all particle traces.}
\end{figure}

\textbf{Results:} \quad A time-lapse depiction of this experiment is presented in Figure~\ref{fig:links-walking}.  We recorded the roll orientation of the robot base and labeled a configuration a fall if the roll exceeded $\frac{\pi}{2}$ radians from vertical.  We observed that \software{Links} completed the 20s walk \emph{in situ} without falling when no particle traces exhibited a fall.  When the robots fell in some traces, \software{Links} walked for 10s \emph{in situ} before falling.  When the models in all traces fell (after the first step), \software{Links} fell on its first step \emph{in situ} as well (see Figure~\ref{fig:links-data}).  While a simulation of the robot from modeling and estimates might not have exhibited the robot's \emph{in situ} behavior for a given policy, the aggregate behavior over all particle traces matched the \emph{in situ} performance well (see Figure~\ref{fig:links-time}).

\section{Discussion} 
We have been able to locate seemingly hard to locate grazing bifurcations (i.e., ones that lie in small volumes in state space) for high dimensional systems with very few particle traces. Since each particle trace is completely independent, particle traces 
can be generated in an ``embarassingly parallel'' manner. So not only is the particle trace approach versatile and simple to implement, it can be quite fast given sufficient computational resources.

We believe the following questions now require much deeper investigation: 
\1 What scenarios can be constructed for which grazing bifurcations are 
computationally demanding to find (much like the ``bugtrap'' scenarios for
motion planning)?; \2 What other dynamic scenarios can statistical 
ensembles of physically simulated robots reliably characterize (and where
will such simulations fail to characterize behavior)?; \3 Since simulations
are capable of generating huge quantities of data, how can such state space
telemetry data be efficiently ``mined'' to locate clusters of similar
high-level behavior? 

\bibliographystyle{plain}
\bibliography{../../../Bibliography/paper}

\end{document}